\documentclass[letterpaper, 10 pt, conference]{ieeeconf}  

\IEEEoverridecommandlockouts                              

\usepackage{graphics} 
\usepackage{epsfig} 
\usepackage{amsmath} 
\usepackage{amssymb}  
\usepackage{ upgreek }
\usepackage[linesnumbered,ruled]{algorithm2e}

\newcommand{\bm}[1]{\boldsymbol{\mathbf{#1}}}

\newcommand{\X}{\bm{X}}

\newcommand{\p}{{\bm{p}}}

\newcommand{\K}{\bm{K}^\mathcal{P}}
\newcommand{\fe}{\bm{f}^e}

\newcommand{\spd}{\bm{\mathcal{S}}_{++}^{m}}
\newcommand{\sym}{\bm{Sym}^{m}}
\newcommand{\trsp}{\mathsf{T}}
\newcommand{\etal}{\MakeLowercase{\textit{et al.\ }}}
\usepackage{xcolor}

\usepackage{graphicx}
\usepackage{float}
\usepackage{amsmath}

\usepackage{todonotes}

\usepackage{subfigure}
\usepackage{algpseudocode}
\usepackage{epstopdf}
\usepackage{multicol}
\usepackage[leftcaption]{sidecap}
\usepackage{booktabs}

\usepackage{amssymb}
\usepackage{arydshln}

\title{\LARGE \bf
	Geometry-aware Dynamic Movement Primitives
}

\author{Fares J. Abu-Dakka and Ville Kyrki
	\thanks{All authors are with Intelligent Robotics Group, Department of Electrical Engineering and Automation (EEA), Aalto University,	Espoo, Finland.
		{\tt\small fares.abu-dakka, ville.kyrki@aalto.fi}}%
}

\begin{document}

	\maketitle
	\thispagestyle{empty}
	\pagestyle{empty}

	\begin{abstract}
		
	  In many robot control problems, factors such as stiffness and damping matrices and manipulability ellipsoids are naturally represented as symmetric positive definite (SPD) matrices, which capture the specific geometric characteristics of those factors. Typical learned skill models such as dynamic movement primitives (DMPs) can not, however, be directly employed with quantities expressed as SPD matrices as they are limited to data in Euclidean space.
      In this paper, we propose a novel and mathematically principled framework that uses Riemannian metrics to reformulate DMPs such that the resulting formulation can operate with SPD data in the SPD manifold. Evaluation of the approach demonstrates that beneficial properties of DMPs such as change of the goal during operation apply also to the proposed formulation. 
	\end{abstract}

	\section{Introduction}
	
	Day by day realistic robotic applications are bringing robots into human environments such as houses, hospitals, and museums where they are expected to assist us in our daily life tasks. Such human-inhabited environments are highly unstructured, dynamic and uncertain, making hard-coding the environments and related skills infeasible.

        In this context, human expertise can be exploited to teach robots how to perform such tasks by transferring human skills to robots \cite{schaal1999}. Learning-from-human-demonstrations (LfD) has been widely studied as a convenient way to transfer human skills to robots. This learning approach is aimed at extracting relevant motion patterns from human demonstrations and subsequently applying these patterns to different situations. In the past decades, several LfD based approaches have been developed such as: dynamic movement primitives (DMP) \cite{ijspeert2013, AbuDakka2015}, probabilistic movement primitives (ProMP) \cite{paraschos2013}, Gaussian mixture models (GMM) along with Gaussian mixture regression (GMR) \cite{calinon2014}, and more recently, kernelized movement primitives (KMP) \cite{huang2019,huang2019b}. DMPs have several beneficial properties such as robustness against perturbations and ability to adapt to new requirements such as a new goal.
        
        However, many tasks in those environments require variable impedance \cite{ikeura1995,tsumugiwa2002,AbuDakka2018} or high manipulability \cite{Guilamo2006,Vahrenkamp2012,Lee2016,Rozo2017}), the parameters of which are encapsulated in symmetric positive definite (SPD) matrices. Because of the structure of the manifold of SPD matrices, standard LfD approaches such as DMPs can not be directly used as they rely on Euclidean parametrization of the space.

        In this paper, we a novel formulation for DMPs using Riemannian metrics such that the resulting formulation can operate with SPD data. This allows variable SPD quantities to be modeled while retaining the useful properties of standard DMPs. The main contributions are
	\begin{enumerate}
		\item A novel and mathematically principled framework for reformulating DMPs using Riemanian metrics, in order to learn and reproduce SPD-matrices-based robot skills.
		\item Reformulating standard DMP goal switching to be able to handle SPD-matrix-based robot skills.
	\end{enumerate}
        
        This extension of DMPs to Riemannian manifolds allows the generation of smooth trajectories for data that do not belong to the Euclidean space.
        
        The work is inspired by quaternion and rotation matrix based formulations of DMPs \cite{AbuDakka2015,Ude2014} which target specifically the problem of parametrizing the space of orientations $SO(3)$.
        Compared to the tensor-based formulation of GMM and GMR on Riemannian manifold of SPD matrices \cite{Jaquier2017} (demonstrated for manipulability transfer in \cite{Rozo2017}), the proposed approach allows adapting the SPD profile on-line to a new goal, similar to the difference between standard GMM/GMR and DMPs for Euclidean quantities.

	This paper is organized as follows: We begin by providing background about standard DMPs (Section \ref{sec:dmps}) and Riemannian manifold of SPD matrices (Section \ref{sec:riemannian}). Afterwards, we exploit Riemannian manifold to derive the new formulation of DMPs (Section~\ref{sec:dmpFormulation}) followed by goal switching formulation (Section~\ref{sec:goalSwitching}). Subsequently, we evaluate our approach through several examples (Section~\ref{sec:exper}). The work is concluded in Section~\ref{sec:concl}.
	
	\section{Background}
	\label{sec:background}
	
	In this scope we introduce a brief introduction to standard DMPs and Riemannian manifold of SPD matrices.
	
	\subsection{Dynamic Movement Primitives}
	\label{sec:dmps}
	The basic idea of DMPs is to model movements by a system of differential equations that ensure some desired behavior, e.g. convergence to the specified attractor point \cite{Schaal2007,ijspeert2013,AbuDakka2015}. A DMP for a single degree of freedom trajectory $y$ is defined by the following set of nonlinear differential equations \cite{ijspeert2013}
	\begin{eqnarray}
		\tau\dot{z} & = & \alpha_z(\beta_z(g - y) - z) + f(x), \label{eq:DMP_accel} \\
		\tau\dot{y} & = & z, \label{eq:DMP_velocity}\\
		\tau\dot{x} & = & -\alpha_xx, \label{eq:DMP_phase}
	\end{eqnarray}
	where $z$ is the scaled velocity, $x$ is the phase variable to avoid explicit time dependency and $x(0)=1$, $\alpha_z$ and $\beta_z$ define the behavior of the 2nd--order system, $g$ is the goal of the movement, and $f(x)$ is a nonlinear forcing term that provides a modeling of complex trajectories. Choosing a time constant $\tau > 0$ along with $\alpha_z=4\beta_z$ and $\alpha_x>0$ will make the linear part of \eqref{eq:DMP_accel} and \eqref{eq:DMP_velocity} critically damped, which insures the convergence of $y$ and $z$ to a unique attractor point at $y=g$ and $z=0$ \cite{ijspeert2013}. $f(x)$ is defined as a linear combination of $N$ nonlinear radial basis functions, which enables the robot to follow any smooth trajectory from the initial position $y_0$ to the final configuration $g$
	\begin{equation}
		f(x) = \frac{\sum_{i=1}^Nw_i\Psi_i(x)}{\sum_{i=1}^N\Psi_i(x)}x, 
		\label{eq:dmp_fx}
	\end{equation}
	\begin{equation}
		\Psi_i(x) = \exp\left(-h_i\left(x-c_i\right)^2\right),
		\label{eq:dmp_psi}
	\end{equation}
	where $c_i$ are the centers of Gaussians distributed along the phase of the movement and $h_i$ their widths. For a given $N$ and setting $\tau$ equal to the total duration of the desired movement, we can define $c_i=\text{exp}\big(-\alpha_x\frac{i-1}{N-1}\big)$, $h_i=\frac{1}{(c_{i+1} - c_i)^2}$ and $h_N=h_{N-1}$ where $i = 1,\dots,N$. For each DoF, the weights $w_i$ should be adjusted from the measured data so that the desired behavior is achieved. For controlling a robotic system with more than one DoF, we represent the movement of every DoF with its own equation system \eqref{eq:DMP_accel}--\eqref{eq:DMP_velocity}, but with the common phase \eqref{eq:DMP_phase} to synchronize them.
	
	\subsection{Riemannian manifold of SPD matrices}
	\label{sec:riemannian}
	
	Let us define $\spd$ as the set of $m\times m$ SPD matrices which cannot be considered as a vector space since it is not closed under addition and scalar product \cite{Pennec2006}, and thus the use of classical Euclidean space methods for treating and analyzing these matrices is inadequate. A compelling solution is to incorporate these matrices with a Riemannian metric, which allows the set of SPD matrices to form a Riemannian manifold \cite{Pennec2006}. Such a metric defines the geodesics, that is, the minimum length curves between two points on the manifold.
	
	A Riemannian manifold $\mathcal{M}$ is a topological space, each point of which locally resembles a Euclidean space. For each point ${\p\in\mathcal{M}}$, there exists a tangent space ${\mathcal{T}_\p\mathcal{M}}$ which corresponds to the space of symmetric matrices for the case of the SPD manifold. The metric in the tangent space is flat, which allows the use of classical arithmetic tools.
        Note that the space of $\spd$ can be represented as the interior of a convex cone embedded in its tangent space of symmetric $m \times m$ matrices $\sym$.

        To operate on the tangent spaces, a mapping system is required to switch between ${\mathcal{T}_\p\mathcal{M}}$ and $\mathcal{M}$. The two mapping operators are known as exponential and logarithmic maps:
	
	\noindent
	\textit{The logarithmic map } ${\text{Log}_{\bm{\Gamma}}(\bm{Q})\!:\!\mathcal{M}\mapsto\mathcal{T}_{\bm{\Gamma}}\mathcal{M}}$ is a function that maps a point in the manifold $\bm{Q}\in \mathcal{M}$ to a point in the tangent space$\bm{\Delta} \in \mathcal{T}_{\bm{\Gamma}}\mathcal{M}$.
	\begin{equation}
		\text{Log}_{\bm{\Gamma}}(\bm{Q}) = 	\bm{\Gamma}^\frac{1}{2}\text{logm}\Big(\bm{\Gamma}^{-\frac{1}{2}}\bm{Q}\bm{\Gamma}^{-\frac{1}{2}}\Big)\bm{\Gamma}^\frac{1}{2},
		\label{eq:log}
	\end{equation}

	\noindent
	\textit{The exponential map} $\text{Exp}_{\bm{\Gamma}}(\bm{\Delta})\!:\!{\mathcal{T}_{\bm{\Gamma}}\mathcal{M}\mapsto\mathcal{M}}$ is a function that maps a point $\bm{\Delta}\in \mathcal{T}_{\bm{\Gamma}}\mathcal{M}$ to a point ${\bm{Q}\in\mathcal{M}}$, so that it lies on the geodesic starting from ${\bm{\Gamma}\in \spd}$ in the direction of $\bm{\Delta}$.
	\begin{equation}
		\text{Exp}_{\bm{\Gamma}}(\bm{\Delta}) = \bm{\Gamma}^\frac{1}{2}\text{expm}\Big(\bm{\Gamma}^{-\frac{1}{2}}\bm{\Delta}\bm{\Gamma}^{-\frac{1}{2}}\Big)\bm{\Gamma}^\frac{1}{2},\label{eq:exp}
	\end{equation}
	where $\text{logm}(\cdot)$ and $\text{expm}(\cdot)$ are the matrix logarithm and exponential functions.
	
	Moving elements between different tangent spaces is performed by the parallel transport operator \cite{Sra2015,Yair2019}. The parallel transport ${\mathbb{B}_{\bm{\Gamma}\mapsto\bm{Q}}(\bm{V}):\mathcal{T}_{\bm{\Gamma}}\mathcal{M}\mapsto\mathcal{T}_{\bm{Q}}\mathcal{M}}$ is a function that transports $\bm{V} \in \mathcal{T}_{\bm{\Gamma}}\mathcal{M}$ to $\mathcal{T}_{\bm{Q}}\mathcal{M}$ over the geodesic from $\bm{\Gamma}$ to $\bm{Q}$ is given by
	\begin{equation}
		\mathbb{B}_{\bm{\Gamma}\mapsto\bm{Q}} (\bm{V}) = \bm{C}_{\bm{\Gamma}\mapsto\bm{Q}} \; \bm{V} \; \bm{C}_{\bm{\Gamma}\mapsto\bm{Q}}^\trsp,
		\label{eq:paralellTrans}
	\end{equation} 
	where ${\bm{C}_{\bm{\Gamma}\mapsto\bm{Q}} = (\bm{Q}\bm{\Gamma}^{-1})^{\frac{1}{2}}}$. Equation \eqref{eq:paralellTrans} has been proved to be computationally efficient \cite{Yair2019}. This transporter is exploited whenever it is required to transport SPD matrices along geodesics in a nonlinear manifold.
	
	In this paper, we exploit the Riemannian manifold to reformulate DMPs to be capable of encoding and reproducing SPD-matrices-based robot skills.

	\section{Proposed Approach}
	\label{sec:proposed}
	In this section, we provide a complete formulation for DMPs in order to learn and reproduce SPD-matrices-based robot skills. For the sake of simplicity let us first recall the re-interpretation of basic standard operations in a Riemannian manifold (Table \ref{tab:basicOperation}). Define $\bm{A}, \bm{B}\in \mathcal{M}$ and $\bm{a},\bm{b}\in \mathbb{R}^n$.

	\begin{table}[!ht]
		\caption{Re-interpretation of basic standard operations in a Riemannian manifold \cite{Pennec2006}.}
		\label{tab:basicOperation}
		\begin{center}\vspace{-0.4cm}
			\begin{tabular}{lcc}\cline{2-3}
								& Euclidean space       & Riemannian manifold       \\ \hline
				Subtraction		& $\overrightarrow{\bm{a}\bm{b}}=\bm{b}-\bm{a}$	& $\overrightarrow{\bm{A}\bm{B}}=\text{Log}_{\bm{A}}(\bm{B})$             \\
				Addition  		& $\bm{b} = \bm{a}+\overrightarrow{\bm{a}\bm{b}}$ & $\bm{B}=\text{Exp}_{\bm{A}}(\overrightarrow{\bm{A}\bm{B}})$             \\
				Distance		& $\text{dist}(\bm{a},\bm{b})=\parallel \bm{b}-\bm{a}\parallel$ & $\text{dist}(\bm{A},\bm{B})=\parallel\overrightarrow{\bm{A}\bm{B}}\parallel_{\bm{A}}$              \\
				Interpolation	& $\bm{a}(t)=\bm{a}_1+t \overrightarrow{\bm{a}_1\bm{a}_2}$ & $\bm{A}(t)=\text{Exp}_{\bm{A}_1}(t\overrightarrow{\bm{A}_1\bm{A}_2})$             \\ \hline
			\end{tabular}
		\end{center}\vspace{-0.5cm}
	\end{table}
	
	\subsection{Geometry-aware DMPs Formulation}
	\label{sec:dmpFormulation}
	
	Define a variable $\X \in \spd$ as an arbitrary SPD matrix and $\bm{\Xi} = \{t_l, \X_l\}_{l=1}^T$ as the set of SPD matrices in one demonstration. In order to prepare the demonstration data for DMP, its 1st- and 2nd-time derivatives are needed. The 1st-time derivative is computed as follows
	\begin{equation}
		\bm{\Sigma}\equiv\bm{\dot{\Xi}} = (\text{Log}_{\X_{l-1}}(\X_l))/dt,
	\end{equation}
	where each $d\bm{\Xi}_t$ belongs to the corresponding tangent space $\mathcal{T}_{\X_{l-1}}\mathcal{M}$.
        Afterwards, we use \eqref{eq:paralellTrans} to move all $d\bm{\Xi}_l$ to a common/shared arbitrary tangent space, e.g. the tangent space of the first SPD data $\mathcal{T}_{\X_{1}}\mathcal{M}$. The tangent space $\mathcal{T}_{\X_{1}}\mathcal{M}$ corresponds to $\sym$, which allows the use of classical arithmetic tools as mentioned in section \ref{sec:riemannian}. To avoid replicating information due to symmetry, we propose to reduce the space dimensionality of the data in the tangent space to $n = {m+m(m-1)/2}$ using Mandel's representation. 
	\begin{equation}
		vec(\bm{\Sigma}) = \bm{\sigma},\quad \bm{\sigma}\in \mathbb{R}^n,
	\end{equation}
	where $vec(\cdot)$ is a function that transforms a symmetric matrix into a vector using Mandel's notation. E.g. a vectorization of a $2\times 2$ symmetric matrix is
	\begin{equation}
		vec\Bigg(	\begin{pmatrix} 		a & b \\		b & d 	\end{pmatrix}\Bigg) = 
		\begin{pmatrix} 		a \\		d \\ \sqrt{2}b 	\end{pmatrix}.
	\end{equation}
	 
	Now, the 2nd-derivatives $\dot{\bm{\Sigma}}$ can be computed straight forward using standard Euclidean tools and its vectorization is denoted as $\dot{\bm{\sigma}}$. Having all necessary data $\{t_l,\X_l,\bm{\sigma}_l,\bm{\dot{\sigma}}_l\}_{l=1}^T$, we transform the standard DMP system \eqref{eq:DMP_accel}--\eqref{eq:DMP_velocity} into a geometry-aware form as follows
	\begin{align}
		\begin{split}
			\tau\dot{\bm{\sigma}} & =  \alpha_z(\beta_z vec(\mathbb{B}_{\X_l\mapsto\X_1} ({\text{Log}_{\X_l}(\X_g)})) - \bm{\sigma}) \\
			&\quad + \bm{\mathcal{F}}(x), \label{eq:DMP_SPD_accel}
		\end{split}\\
		\tau\dot{\bm{\xi}} & =  \bm{\sigma}, \label{eq:DMP_SPD_velocity}
	\end{align}
	where $\bm{\xi}$ is the vectorization of $\bm{\Xi}$. $\X_g\in \spd$ represents the goal SPD matrix. $vec(\mathbb{B}_{\X_l\mapsto\X_1} ({\text{Log}_{\X_l}(\X_g)}))$ is the vectorization of the transported symmetric matrix ${\text{Log}_{\X_l}(\X_g)}$ over the geodesic from $\X_l$ to $\X_1$. The forcing term $\bm{\mathcal{F}}(x)$ can be recalculated as
	\begin{align}
		\begin{split}
			&\bm{\mathcal{F}}(x) = \frac{\sum_{i=1}^N\bm{\mathcal{W}}_i\Psi_i(x_l)}{\sum_{i=1}^N\Psi_i(x_l)}x_l = \\
			&\tau\dot{\bm{\sigma}}-\alpha_z(\beta_z vec(\mathbb{B}_{\X_l\mapsto\X_1} ({\text{Log}_{\X_l}(\X_g)})) - \bm{\sigma}), 
		\end{split}
		\label{eq:dmp_spd_fx}
	\end{align}
	where the phase $x_l=x(t_l)=\text{exp}(-\frac{\alpha_x}{\tau}t_l)$. Using \eqref{eq:dmp_spd_fx}, the weights $\bm{\mathcal{W}}_l\in \mathbb{R}^n$ can be estimated by encoding any sampled SPD-matrices-based robot skills.
	
	In the reproduction, equation \eqref{eq:DMP_SPD_velocity} is integrated as follows
	\begin{equation}
		\bm{\hat{X}}(t+\delta t)=\text{Exp}_{\X(t)}\Bigg(\frac{\mathbb{B}_{\X_1\mapsto\X(t)} (mat(\bm{\sigma}(t)))}{\tau}\delta t\Bigg).
	\end{equation}
	where the function $mat(\cdot)$ is the inverse of $vec(\cdot)$ and denotes to the matricization using Mandel's notation. $\bm{\hat{X}}\in \spd$ represents the new SPD-matrices-based robot skills.
	
	\subsection{Goal switching}
	\label{sec:goalSwitching}
	
	In the standard DMP formulation, in case of sudden goal switching (e.g. based on external sensory information) during the execution, Ijspeert \etal \cite{ijspeert2013} proposed to add an additional equation to the dynamic system \eqref{eq:DMP_accel}--\eqref{eq:DMP_velocity} in order to smoothly change the goal $g$ in \eqref{eq:DMP_accel} to a new goal $g_{new}$ as
	\begin{equation}
		\tau \dot{g} = \alpha_g(g_{new}-g),
	\end{equation}
	where $\alpha_g$ is a constant. This equation transforms $g$ from being a constant to a continuous variable. Analogously, SPD-based DMP can switch the goal using
	\begin{equation}
		\tau \bm{\dot{g}} = \alpha_g\text{Log}_{\bm{g}}(\bm{g}_{new}).
	\end{equation}
	so $\bm{g}$ now is updated continually. 
	
	\section{Experiments}
	\label{sec:exper}
	We evaluated the proposed imitation learning framework using simulated data. All algorithms have been implemented in MATLAB\textsuperscript{\textregistered} using a workstation running Ubuntu 16.04 LTS with Intel Core i9--8950HK CPU @ 2.90GHz $\times$ 12, 16 GB of RAM.
	
	Four different experiments were carried out to evaluate the proposed framework:
	\begin{itemize}
		\item Learning and reproducing full stiffness matrix profiles with a 2-DoF virtual-mass spring-damper system (MSD).
		\item Goal switching applied to full stiffness matrix profiles.
		\item Comparison of the resulting SPD profile between the proposed DMP and GMM/GMR proposed in \cite{Jaquier2017}.
	\end{itemize}

	\subsection{Learning Variable Impedance Skills}
	\label{sec:LearVarImp}
	
	\begin{figure}[!t]
		\centering
		\includegraphics[width=1\linewidth]{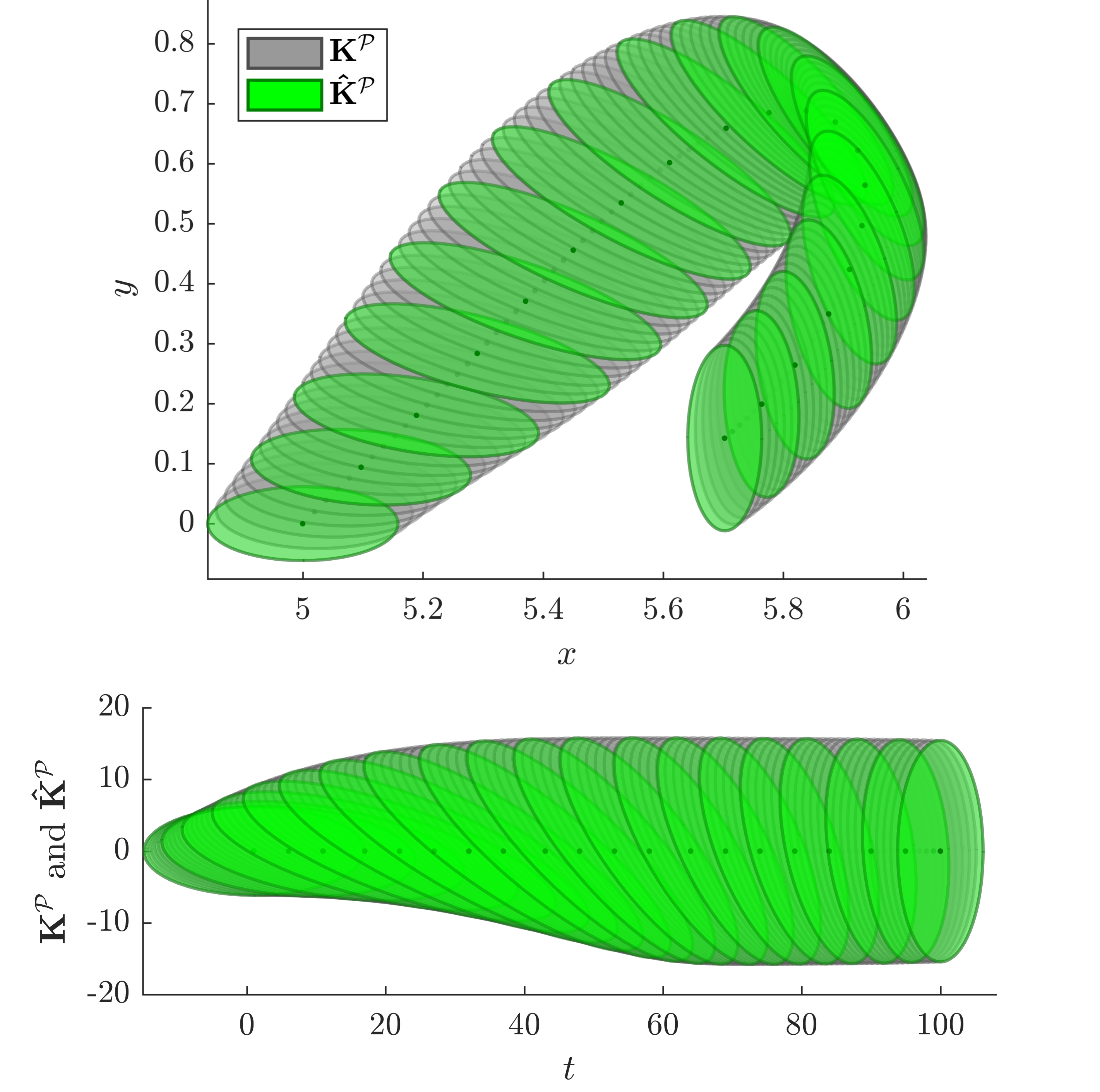} 
		\vspace{-0.7cm}
		\caption{Learning and reproduction of variable stiffness profile for MSD system. \emph{Top:} The reproduced stiffness ellipsoids in different time steps (in green) coinciding the demonstration (in gray) over the Cartesian trajectory of the MSD. \emph{Bottom:} The reproduced stiffness ellipsoids in different time steps (in green) coinciding the ground truth (in gray) over time.}
		\vspace{-0.5cm}
		\label{fig:stifEllipsoids}
	\end{figure}

	In this scope, we propose to use a simulation of MSD to evaluate our geometry-aware DMPs for learning and reproducing variable impedance\footnote{Here we refer to variable impedance as variable stiffness profiles. Nevertheless, the proposed approach can also be used to learn variable damping controllers as well as any SPD-matrix-based robot skills.} skills. Note that stiffness matrices $\K$ belong to the space of $\spd$. The MSD system starts from an initial, horizontally-aligned, stiffness ellipsoid $\K$ at rest position. Afterwards, external forces $\fe$ are applied to stimulate the MSD system. During the stimulation, $\K$ is rotating through ${\mathbf{R}^\trsp\K\mathbf{R}}$ ($\mathbf{R}$ is a rotation matrix) until it ends up with a vertically-aligned ellipsoid as shown in Fig.~\ref{fig:stifEllipsoids} in gray. This demonstration then is encoded using \eqref{eq:DMP_SPD_accel}--\eqref{eq:DMP_SPD_velocity} to reproduce the ellipsoids in green $\bm{\hat{K}}^\mathcal{P}$. From the figure, we can see the match between the results of the SPD-based DMPs and the demonstration.

\begin{figure}[t]
		\centering
		\includegraphics[width=\linewidth]{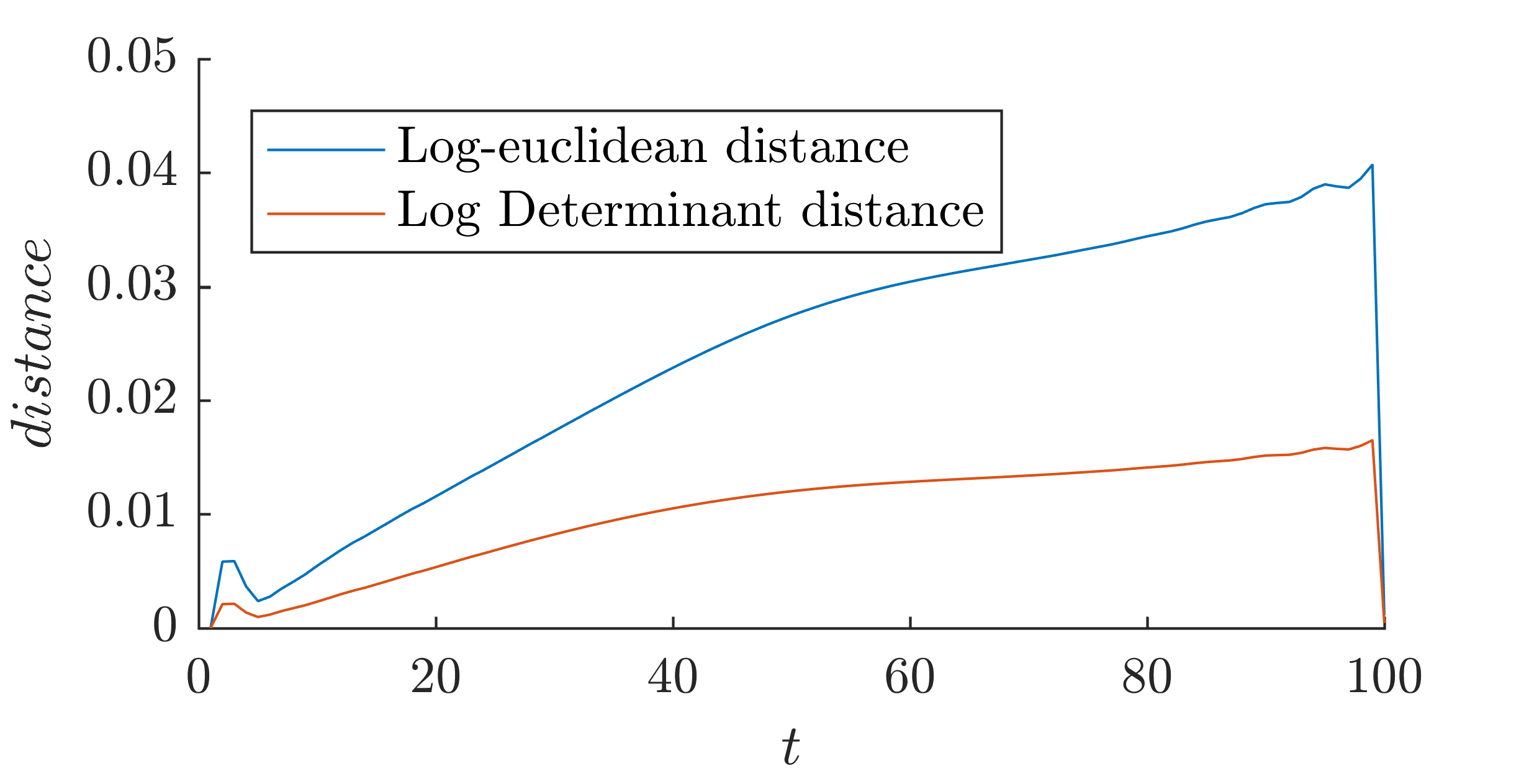} 
		\vspace{-0.7cm}
		\caption{The distance between the resulting stiffness profile from the proposed DMPs and the ground truth using different metrics.}
		\label{fig:stifDist}
	\end{figure}

	Figure~\ref{fig:stifDist} tests the accuracy of the proposed SPD-based DMP by calculating the distance between the resulting SPD profile and the demonstration one. Here we used two distance metric systems: (\textit{i}) Log-Euclidean distance \cite{Arsigny2006}. (\textit{i}) Jensen-Bregman Log-Determinant distance \cite{Cherian2011}. 
	
	\subsection{Goal Switching}
	\label{sec:goalSwitchingEXP}
	
	\begin{figure}[t]
		\centering
		\includegraphics[width=\linewidth]{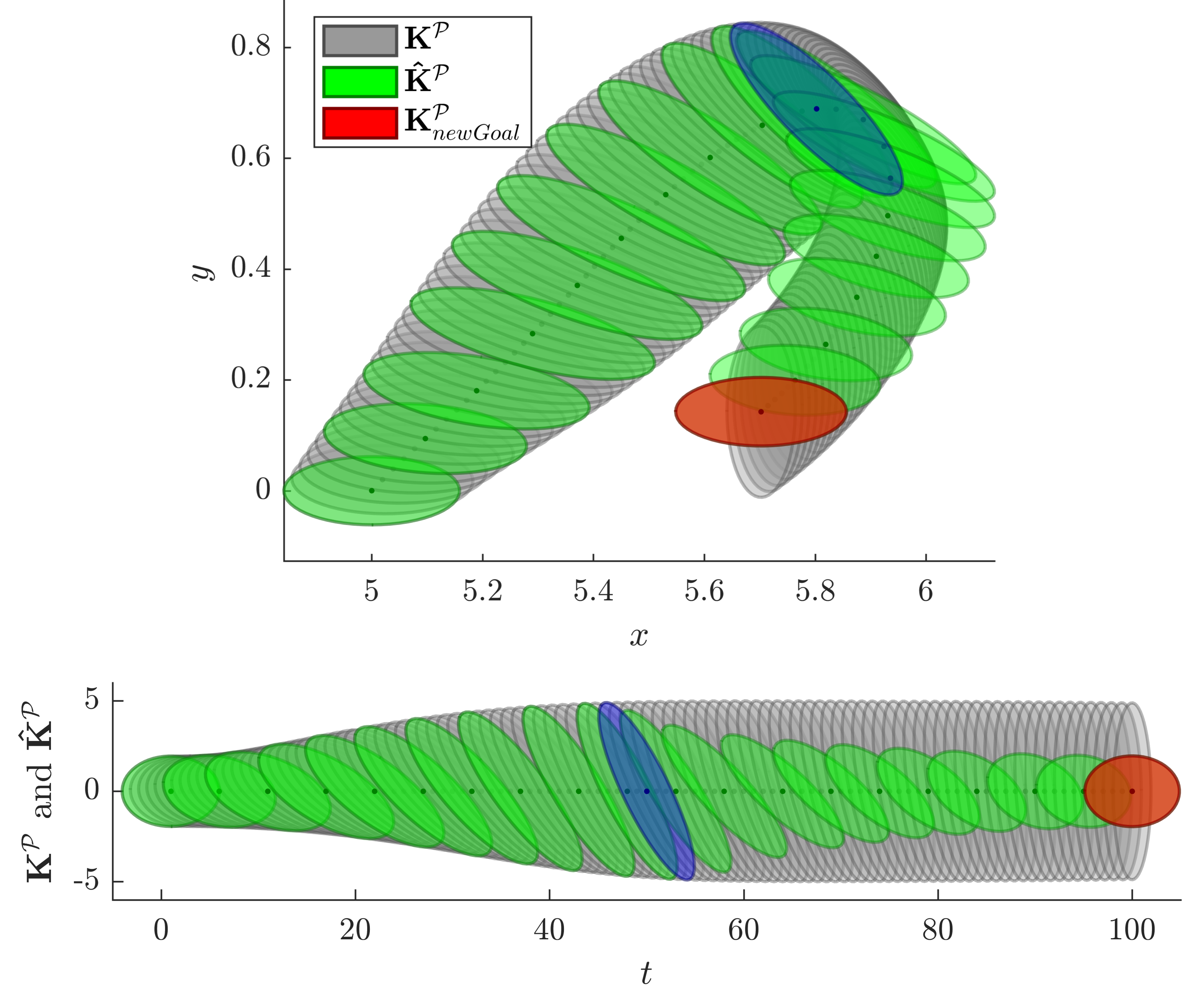} 
		\vspace{-0.7cm}
		\caption{Response of the proposed DMPs to the switching of stiffness goal during the movement over Cartesian trajectory (\emph{Top}) and over Time (\emph{Bottom}). The stiffness goal was changed at the middle of the movement (blue ellipsoid). The new goal (in red) is the original ellipsoid but rotated by 90 degrees.}
		\vspace{-0.5cm}
		\label{fig:stifElliGoal}
	\end{figure}
	
	In this simulation we used the same MSD setup introduced in section \ref{sec:LearVarImp}. However, here we are about to test the response of the proposed SPD-based DMP to sudden goal changing during the execution. At the middle of DMP execution we changed the goal by rotating it 90 degrees to be horizontally-aligned (red ellipsoid in Fig.~\ref{fig:stifElliGoal}) instead of being vertically-aligned (in gray). 
	
\begin{figure}[t]
		\centering
		\includegraphics[width=\linewidth]{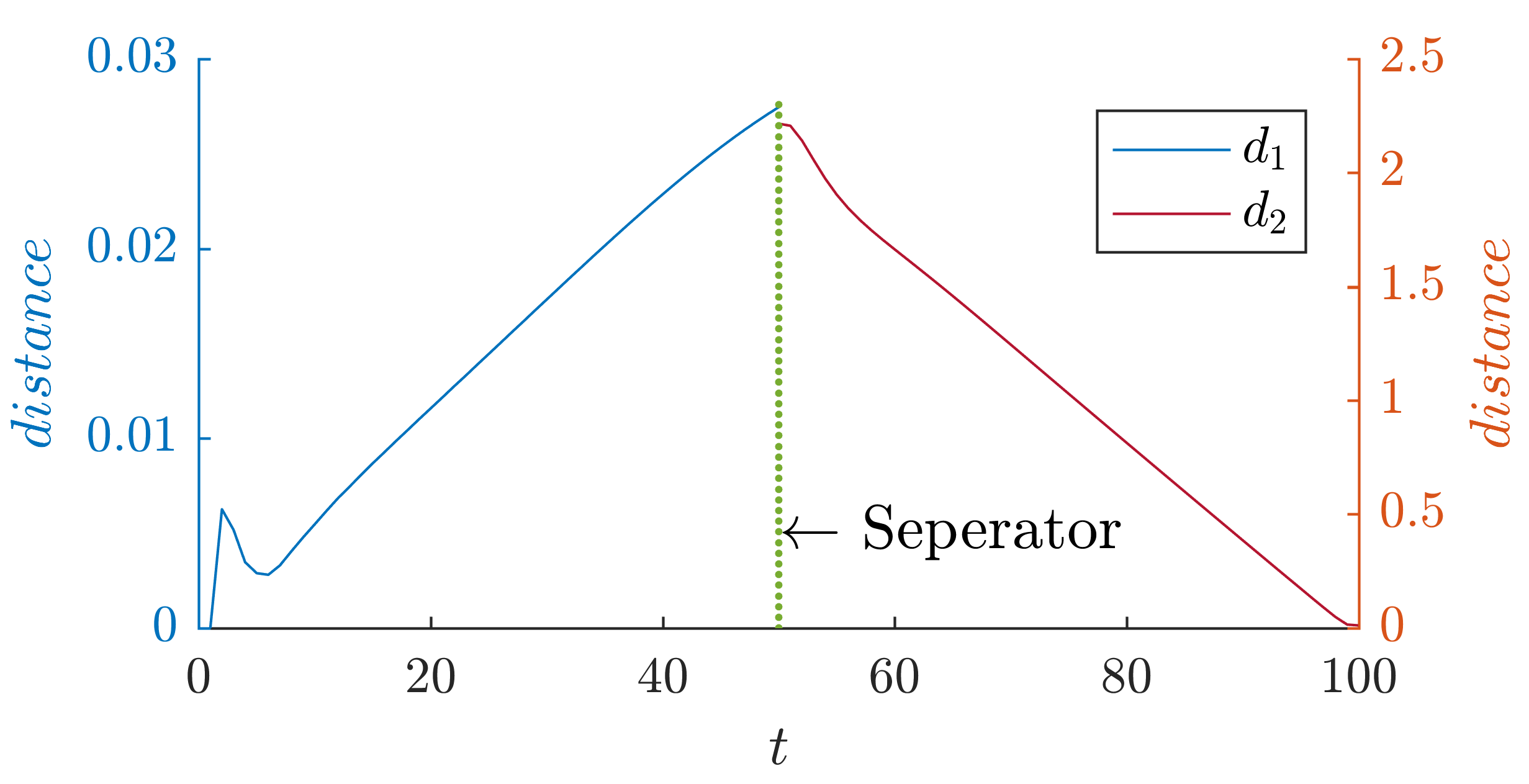} 
		\vspace{-0.7cm}
		\caption{$d_1$ Represents the distance between the demonstrated stiffness profile and the resulting DMP profile just before the goal switching occurs. $d_2$ is the distance between the resulting stiffness from the DMPs and the new goal. $d_2$ shows how the DMP is adapting and approaching to the new goal.}
		\vspace{-0.5cm}
		\label{fig:stifDistGoal}
	\end{figure}

	Figure~\ref{fig:stifElliGoal} shows the smoothness of the adaptation of the stiffness profile (in green) to the new goal (in red). The blue stiffness ellipsoid marks the instant of goal switching. It is clear from the figure that the resulting profile was following the demonstrated one until the blue ellipsoid, then started to adapt to the new goal. More clearification regarding the accuracy of the approach can be seen in Fig.~\ref{fig:stifDistGoal}. The blue part of the figure shows the distance before the occurrence of goal switching. However, the red part shows the distances between the SPD-based DMP results and the new goal. The figure illustrates that the system converges to the new goal. 
	
	\subsection{Comparison with GMM/GMR}
	\label{sec:compare}
	
	In this scope we compare the proposed SPD-based DMP with SPD-based GMM/GMR proposed by \cite{Jaquier2017}. For fair comparison, as DMP is trained using one demonstration, we used also this same one demonstration to train GMM. 
	As number of Gaussian components influence the accuracy of GMM/GMR,  we trained 1-, 4-, 7-, and 10-states GMMs. 
	
\begin{figure}[t]
		\centering
		\includegraphics[width=\linewidth]{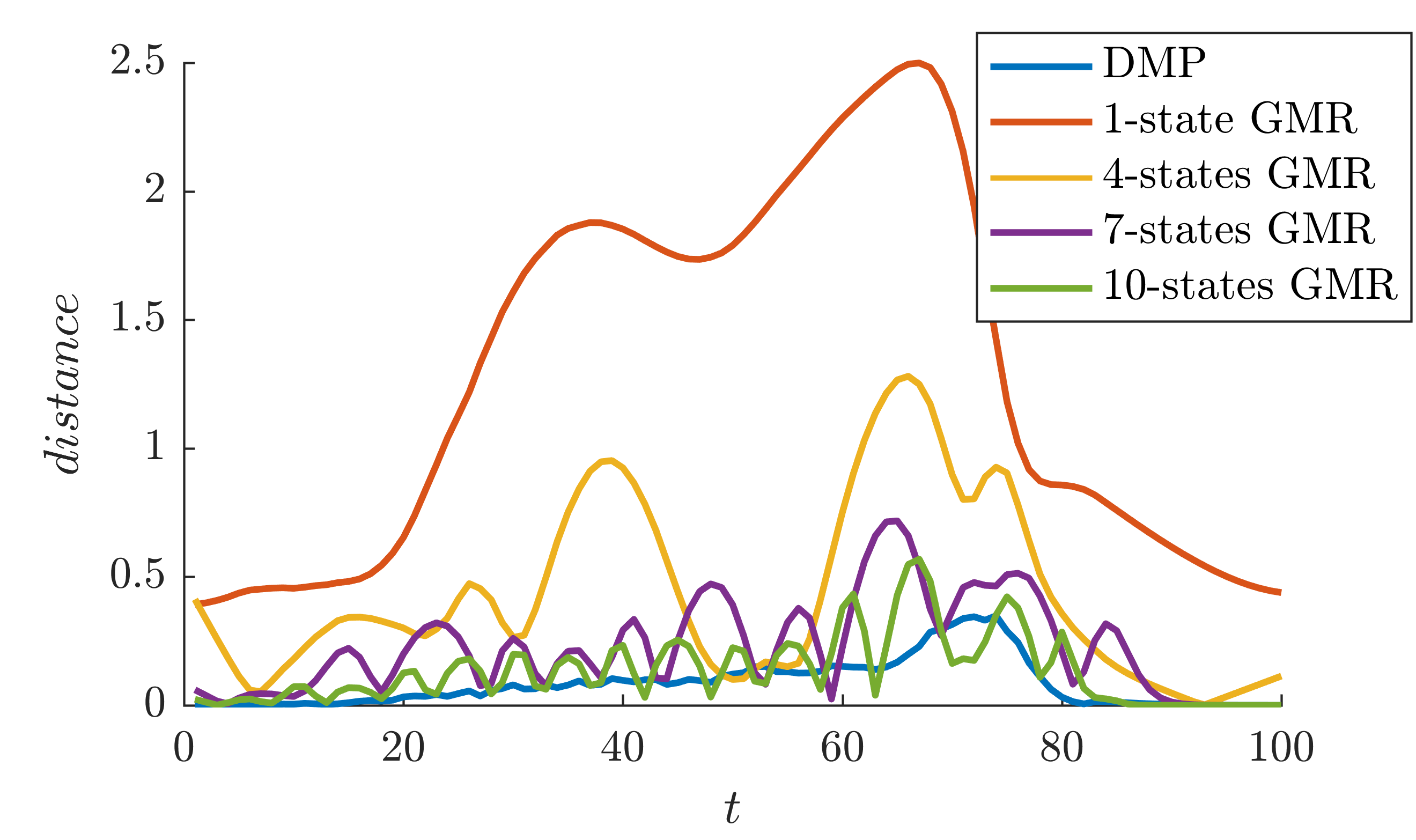} 
		\vspace{-0.7cm}
		\caption{The log-Euclidean distance between the resulting SPD profile from the proposed DMPs and the ground truth using different metrics.}
		\vspace{-0.5cm}
		\label{fig:manDistGMR}
	\end{figure}

	For each GMM model, we calculated the distance error between the SPD profile obtained by GMR and the demonstration. Moreover, the distance error also has been calculated in the case of the proposed SPD-based DMP. Figure~\ref{fig:manDistGMR} shows the resulting distance error in all cases. From the figure, it is clear that the accuracy of GMM/GMR increases when the number of Gaussian components increases. However, increasing Gaussian components leads to a significant increase in the computation time as shown in Table \ref{tab:elapsed_time}, while the proposed SPD-based DMP is significantly faster. Moreover, the GMM/GMR approach would not allow e.g.~goal switching.
	
	\begin{table}[!ht]
		\caption{Comparison of the computational cost between the proposed DMP and GMM/GMR \cite{Jaquier2017}}
		\vspace{-0.4cm}
		\label{tab:elapsed_time}
		\begin{center}
			\begin{tabular}{lcc}
				\cline{2-3}
				& \multicolumn{2}{c}{Time in seconds} \\ \cline{2-3} 
				& Learning       & Reproduction       \\ \hline
				1-state GMM/GMR   & 2.8354         & 1.0788             \\
				4-states GMM/GMR  & 9.9459         & 3.9176             \\
				7-states GMM/GMR  & 16.3209        & 6.1992             \\
				10-states GMM/GMR & 22.2517        & 8.3852             \\ \hline
				Proposed SPD-based DMP  & 0.0941         & 0.4214       \\ \hline
			\end{tabular}
		\end{center}
	\vspace{-0.5cm}
	\end{table}

	\section{CONCLUSIONS}
	\label{sec:concl}
	In this paper we successfully exploited the Riemannian manifold of $\spd$ to derive a new formulation of DMPs capable of direct learning and reproduction of SPD-matrix-based robot skills. This formulation avoids any prior reparametrization of such skills. Moreover, we integrated a new formulation for the goal switching that can deal directly with SPD-matrix-based robot skills.
	
	The algorithm has been extensively validated through multiple simulation examples. Moreover, a comparison with GMM/GMR demonstrates that the proposed approach provides at least similar accuracy with a significantly lower computation cost. 
	
	In the future we propose to integrate our approach with other algorithms, e.g. iterative learning control, in order to not just reproduce SPD-matrix-based skills, but also to adapt to different situations and perform more complex tasks (e.g. forced-based variable impedance control). Moreover, we will work on exploration-based learning methods, which will prove to be crucial when a robot needs to significantly adapt to a new situation, e.g. adapt its stiffness, in order to perform successfully in a large diversity of task situations.
        
	\section*{ACKNOWLEDGMENT}
	This work is supported by CHIST-ERA project IPALM (Academy of Finland decision 326304).

	
	\bibliographystyle{IEEEtran}
	\bibliography{ref}

\end{document}